%% file: 00-main.tex
\documentclass[letterpaper, 10 pt, journal, twoside]{ieeetran}  

\usepackage{graphicx}
\usepackage{footnote} 
\usepackage{hyperref}
\hypersetup{%
    pdfborder = {0 0 0}
}

\usepackage[]{siunitx}
\DeclareSIUnit{\degree}{°}
\DeclareSIUnit{\deg}{deg}
\DeclareSIUnit{\nothing}{\relax}
\DeclareSIUnit\pixel{px}

\usepackage{color}
\usepackage{tabularray}
\usepackage{array}
\usepackage{booktabs}

\usepackage{utfsym} 
\newcommand{\cmark}{\usym{1F5F8}}%

\definecolor{chromeyellow}{rgb}{1.0, 0.65, 0.0}

\definecolor{blue}{rgb}{0.0, 0.0, 0.0}
\newcommand{\blue}[1]{\color{blue}~#1} 

\title{\LARGE \bf A Sim-to-Real Deep Learning-based Framework for Autonomous Nano-drone Racing} 

\definecolor{somegray}{rgb}{0.5, 0.5, 0.5}
\newcommand{\darkgrayed}[1]{\textcolor{somegray}{#1}}
\makeatletter
\newcommand*\titleheader[1]{\gdef\@titleheader{#1}}
\AtBeginDocument{%
  \let\st@red@title\@title
  \def\@title{%
    \vskip-2.0em
    \bgroup\normalfont\large\centering\@titleheader\par\egroup
    \vskip0.0em\st@red@title}
}

\makeatother

\titleheader{\darkgrayed{This paper has been accepted for publication in the IEEE Robotics and Automation Letters (RAL) \\\copyright 2023 IEEE.}}

\author{
Lorenzo Lamberti$^{1}$, Elia Cereda$^{2}$, Gabriele Abbate$^{2}$, Lorenzo Bellone$^{3}$, Victor Javier Kartsch Morinigo$^{4}$,\\
Micha\l{} Barci\'s$^{3}$, Agata Barci\'s$^{3}$, Alessandro Giusti$^{2}$, Francesco Conti$^{1}$, and Daniele Palossi$^{2,4}$%
\thanks{$^{1}$L. Lamberti and F. Conti are with the Department of Electrical, Electronic, and Information Engineering, University of Bologna, Italy,
        {\tt\small lorenzo.lamberti@unibo.it}}%
\thanks{$^{2}$E. Cereda, G. Abbate, A. Giusti, and D. Palossi are with the Dalle Molle Institute for Artificial Intelligence, USI-SUPSI, Switzerland,}%
\thanks{$^{3}$L. Bellone, M. Barci\'s, and A. Barci\'s are with the Autonomous Robotics Research Center, Technology Innovation Institute, UAE.}%
\thanks{$^{4}$V. J. Kartsch Morinigo and D. Palossi are with the Integrated Systems Laboratory, ETH Z\"urich, Switzerland.}
}

\begin{document}

\maketitle
\thispagestyle{empty}
\pagestyle{empty}


\begin{abstract}
Autonomous drone racing competitions are a proxy to improve unmanned aerial vehicles' perception, planning, and control skills. 
The recent emergence of autonomous nano-sized drone racing imposes new challenges, as their $\sim$\SI{10}{\centi\metre} form factor heavily restricts the resources available onboard, including memory, computation, and sensors.
This paper describes the methodology and technical implementation of the system winning the first autonomous nano-drone racing international competition: the ``IMAV 2022 Nanocopter AI Challenge.''
We developed a fully onboard deep learning approach for visual navigation trained only on simulation images to achieve this goal.
Our approach includes a convolutional neural network for obstacle avoidance, a sim-to-real dataset collection procedure, and a navigation policy that we selected, characterized, and adapted through simulation and actual in-field experiments. 
Our system ranked $1^{st}$ among seven competing teams at the competition.
In our best attempt, we scored \SI{115}{\meter} of traveled distance in the allotted 5-minute flight, never crashing while dodging static and dynamic obstacles.
Sharing our knowledge with the research community, we aim to provide a solid groundwork to foster future development in this field.
\end{abstract}

\bstctlcite{IEEEexample:BSTcontrol}

\section*{Supplementary material}
Supplementary video at: \url{http://youtu.be/vHTAwUsj-nk}


\input{01-introduction}
\input{02-related_work}
\input{03-challenge}
\input{04-navigation_policies}
\input{05-cnn_training_deployment}
\input{06-results.tex}

\input{07-conclusion}

\section*{Acknowledgment}
We thank the Autonomous Robotics Research Center of the Technology Innovation Institute, Abu Dhabi, UAE, for granting us access to their indoor flying arena.
We also thank the organizers of the IMAV'22 conference for sharing with us the trajectory recordings.

\bibliographystyle{IEEEtran}

\bibliography{IEEEabrv,\jobname,biblio}

\end{document}

%% file: 01-introduction.tex
\section{Introduction} \label{sec:intro}

Competitions have always been a catalyst for scientific and technological progress. 
As the \textit{space race} was a driver to develop programmable computers and microchips, likewise, in recent years, autonomous drone racing pushed the development of cutting-edge navigation algorithms, including artificial intelligence (AI), running aboard unmanned aerial vehicles (UAVs)~\cite{decroon_alphapilot,scaramuzza_alphapilot, iros2016_winner}.
Competitions act as a proxy to improve UAVs' perception, planning, and control skills: from 2016 to date, the racetrack's average flight speed of autonomous drones increased from 0.6 to \SI{22}{\meter/\second}~\cite{survey_drone_racing}.
Eventually, these advancements positively impact a more comprehensive range of applications where robust, agile, and precise autonomous navigation is crucial, such as rescue missions~\cite{UAV_safety_rescue} and human-robot interaction~\cite{pulp_frontnet}.

\begin{figure}[t]
\centering
\includegraphics[width=1.0\linewidth]{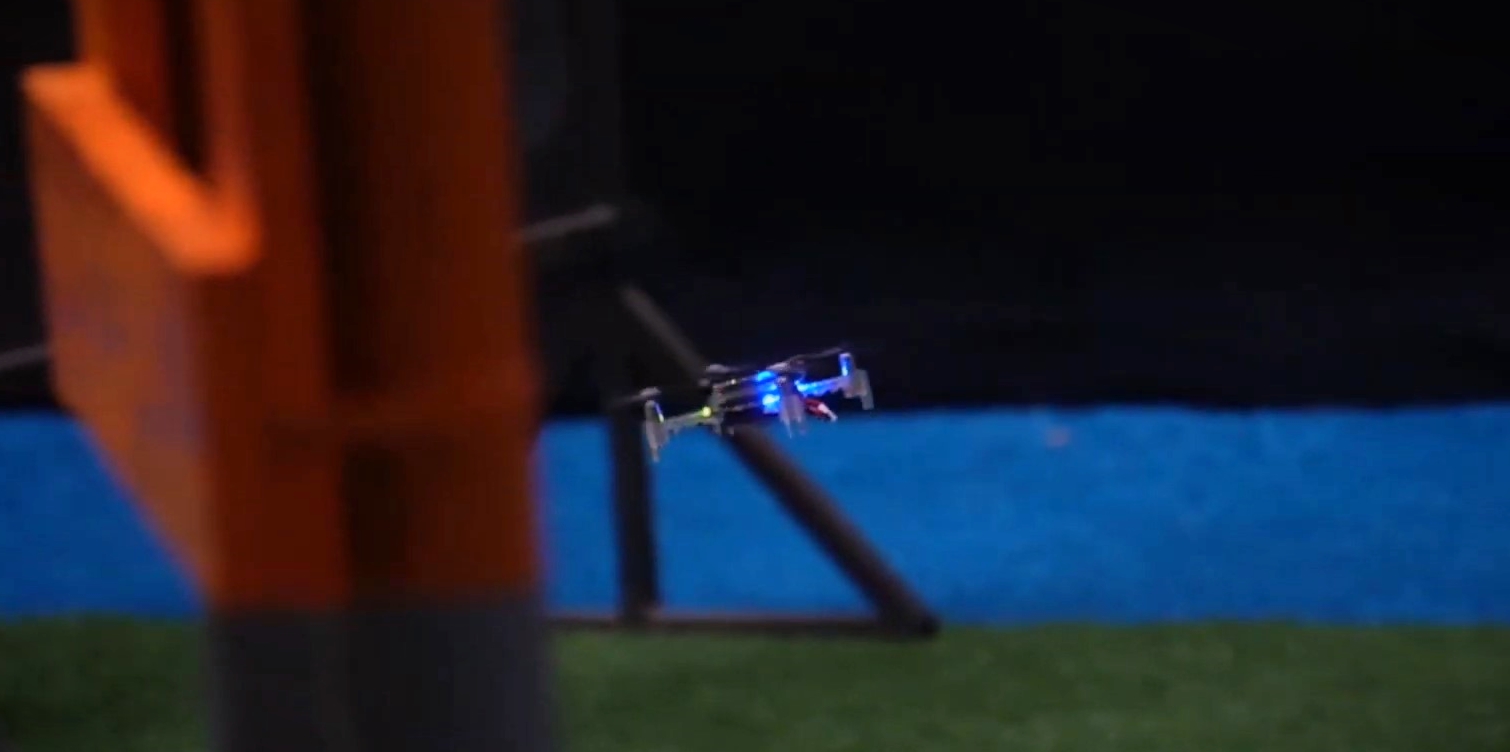}
\caption{Our nano-drone winning the IMAV'22 ``Nanocopter AI Challenge.''}
\label{fig:intro}
\end{figure}

So far, these competitions have focused on micro-sized drones, i.e., $\sim$\SI{30}{\centi\meter}-wide robots capable of hosting powerful processors and rich sensors, \textit{autonomous nano-sized drone racing} constitutes a newborn category employing palm-sized UAVs.
The first competition of this kind was the ``Nanocopter AI Challenge'' hosted at the 13$^{th}$ International Micro Air Vehicle conference (IMAV'22).
This last-born class of robotic competitions poses a new challenge to roboticists due to the small size of nano-drones, i.e., \SI{10}{\centi\meter} in diameter and a few tens of grams in weight.
While nano-drone racing targets tasks similar to major competitions for bigger drones (e.g., AlphaPilot, IROS Drone Racing, etc.), their ultra-tiny form factor allows only minimal onboard resources, i.e., memory, computation, and sensors.
 
\begin{table*}
\label{tab:rw}
\centering
\caption{Literature review for obstacle avoidance (OA) and gate-based navigation (GN) on UAVs.}
\begin{tblr}{
 width = \linewidth,
 colspec = {Q[170] Q[55] Q[76] Q[67] Q[130] Q[100] Q[40] Q[180] Q[60] Q[137]},
 cells = {c},
 hline{1,9} = {-}{0.08em},
 hline{2} = {-}{},
}
\textbf{Work} & \textbf{Size} & \textbf{Onboard} & \textbf{Task} & \textbf{Perception} & \textbf{Algorithm} & \textbf{Data} & \textbf{Compute device} & \textbf{Power} & \textbf{Competition}\\
Wagter \textit{et al.}~\cite{decroon_alphapilot} & Micro & $\cmark$
& GN & Camera & CNN & Real & Jetson AGX Xavier & \SI{30}{\watt} & AlphaPilot'21\\
Foehn \textit{et al.}~\cite{scaramuzza_alphapilot} & Micro & $\cmark$ & GN & Stereo camera & CNN & Real & Jetson AGX Xavier & \SI{30}{\watt} & AlphaPilot'21\\
Kaufmann \textit{et al.}~\cite{kaufmann_iros2018} & Micro & $\cmark$ & GN & Camera & CNN & Sim & Intel UpBoard & \SI{13}{\watt} & IROS'18\\
Jung \textit{et al.}~\cite{iros2016_winner} & Micro & $\cmark$ & GN/OA & Stereo camera & CV & None & Jetson TK1s & \SI{15}{\watt} & IROS'16\\
\blue{Pham \textit{et al.}~\cite{pham_pencilnet_2022}} & \blue{Micro} & \blue{$\cmark$} & \blue{GN} & \blue{Camera} & \blue{CNN} & \blue{Sim} & \blue{Jetson TX2}  & \blue{\SI{15}{\watt}} & ---\\
Niculescu \textit{et al.}~\cite{pulpdronetv2JETCAS} & Nano & $\cmark$ & OA & Camera & CNN & Real & GWT GAP8 & \SI{100}{\milli\watt} & ---\\
\textbf{Ours} & \textbf{Nano} & \textbf{$\cmark$} & \textbf{OA} & \textbf{Camera} & \textbf{CNN} & \textbf{Sim} & \textbf{GWT GAP8} & \textbf{\SI{100}{\milli\watt}} & \textbf{IMAV'22}
\end{tblr}
\end{table*}

In the IMAV'22 competition, all challengers run their navigation algorithms on the same platform: a commercial off-the-shelf (COTS) Crazyflie 2.1 nano-drone equipped with the AI-deck board featuring a GWT GAP8 System-on-Chip (SoC)~\cite{gap8} and a grayscale, low-resolution camera~\cite{palossi2019DCOSS}.
Compared to the typical processors found on micro-sized racing drones~\cite{scaramuzza_alphapilot,decroon_alphapilot}, e.g., Nvidia Jetson Xavier, the GAP8 SoC has more than 1000$\times$ less compute power and memory.
Despite this, the competition encouraged onboard computation by granting a $5\times$ score multiplier while challenging the nano-drone with agile maneuvers to \textit{i}) avoiding static and dynamic obstacles and \textit{ii}) passing through a set of gates in a \textbf{never-seen-before} indoor arena.
In fact, before the competition, no arena map or real-world dataset was released; participants could only access a photorealistic simulator.

\textbf{This paper's main contribution is a thorough analysis and description of the strategy, methodology, and technical implementation we employed to win the IMAV'22 competition: a fully-onboard deep learning-based visual navigation framework trained only on simulation data.}
In detail, we present \textit{i}) an exhaustive discussion of our strategy, which accounts for both the competition's guidelines and the nano-drone's limitations; \textit{ii}) a convolutional neural network (CNN) for obstacle avoidance derived from the open-source \textit{PULP-Dronet}~\cite{pulpdronetv2JETCAS} and trained only on simulation images; \textit{iii}) mitigation of the \textit{sim-to-real} gap~\cite{sim_to_real_gap_survey} via aggressive photometric augmentation, label balancing, and comprehensive data generation; \textit{iv}) three alternative \textit{navigation policies}, which we characterize both in simulation and in the field.

Our final system, employing the best-performing navigation policy, ranked first among six contending teams at the IMAV'22 competition.
In our best run, we scored \SI{115}{\meter} of traveled distance in the allotted \SI{5}{\minute}-flight, never crashing, dodging dynamic obstacles, and only using computational resources aboard our nano-drone.
Our result demonstrates the effectiveness of the proposed sim-to-real mitigation strategy, our implementation's robustness, and our rationale's soundness, and by sharing our insights, we aim to provide the research community with a solid groundwork for the evolution of this field.

%% file: 02-related_work.tex
\section{Related work} \label{sec:related}

As the ``Nanocopter AI Challenge'' focuses on obstacle avoidance (OA) and gate-based navigation (GN), in this section, we focus on these two complex tasks surveying the SoA for various class sizes of drones (Tab.~\ref{tab:rw}).

\textbf{Obstacle avoidance:} while racing micro-drones can carry bulky sensors (e.g., Lidars~\cite{iros2016_winner}, stereo cameras~\cite{iros2016_winner, scaramuzza_alphapilot}) and GPUs with a power envelope of up to \SI{30}{\watt} (Tab.~\ref{tab:rw}), nano-UAVs suffer from limited perception capabilities due to their tiny low-power sensors and microcontroller units (MCUs)~\cite{pulpdronetv2JETCAS}.
State-of-the-Art (SoA) perception algorithms, such as simultaneous localization and mapping (SLAM)~\cite{SLAMBench2}, can not run onboard nano-drones due to their steep performance requirements.
Even when run off-board, they suffer from substantial performance degradation as nano-drones employ low-quality sensors~\cite{slam_offboard_crazyflie, dunkley_visual-inertial_2014}.

Lightweight approaches better suited to nano-drones employ different sensors such as Time-of-Flight (ToF) ranging sensors~\cite{mcguire_minimal_2019-1, pulp_ssd, elkunchwar_bio-inspired_2022}.
These approaches can be implemented onboard and provide robust obstacle avoidance, even in unknown environments, with raw sensor readings or minimal onboard processing (e.g., \SI{30}{OP/\second} in~\cite{elkunchwar_bio-inspired_2022}), such as simple bug-inspired lightweight state machines ~\cite{mcguire_minimal_2019-1,pulp_ssd}.
However, in our competition, only a low-resolution monochrome monocular camera was allowed, narrowing the teams' effort only to visual-based approaches.

A lightweight vision-based system for pocket-sized drones was presented by McGuire~et~al.~\cite{mcguire_efficient_2017}, implementing depth estimation with a stereo camera and achieving obstacle avoidance at low speed (\SI{0.3}{\meter/\second}).
The SoA visual-based CNN for autonomous nano-drone navigation is PULP-Dronet~\cite{pulpdronetv2JETCAS}, trained on real-world data to predict a steering angle and a collision probability based on a QVGA monocular image. 
PULP-Dronet runs onboard the GAP8 SoC at \SI{19}{frame/\second}, proving in-field obstacle avoidance capabilities up to a speed of \SI{1.65}{\meter/\second} when coping with a dynamic obstacle.
As a result, at the IMAV competition, 3 of 6 teams, including us, employed this CNN as a starting point to build their visual obstacle avoidance pipeline.
First, we modified the PULP-Dronet reference implementation in its task, i.e., the original CNN predicts a collision probability and a steering angle.
Instead, our model is optimized to predict three collision probabilities by horizontally splitting the input image into three regions.
Then, we introduced a novel training pipeline that exclusively relies on simulation, while the original work uses real-world images based on autonomous driving cars.
Finally, we enriched our simulator with a photometric augmentation pipeline, which increased the generalization capabilities of our model up to a top-scoring in-field performance.
Compared to PULP-Dronet, this work targets more aggressive obstacle avoidance capabilities: up to \SI{2}{\meter/\second} speed, concurrent static and dynamic obstacles, and \SI{5}{\minute} uninterrupted flight.
A thorough discussion on the comparison with the SoA PULP-Dronet is presented in Sec.~\ref{sec:soa_comparison}.

\textbf{Gate-based navigation:} SoA approaches for micro-drone racing competitions tackle trajectory planning and optimization, gate detection, and control~\cite{survey_drone_racing}.
However, their prohibitive complexity prevents implementing them on nano-drones, even without considering the computational budget needed by the obstacle avoidance task.
Time-optimal trajectory optimization relies on model predictive control (MPC) solving linear algebra at high-frequency ($\sim$\SI{100}{\hertz}) and with real-time constraints~\cite{foehn_time-optimal_2021}.
To cope with the MPC complexity, Foehn~\textit{et~al.}~\cite{foehn_time-optimal_2021} exploit an NVIDIA Jetson TX2 GPU, which has $\sim60\times$ higher computation capabilities than GAP8.

For gate detection, the SoA exploits either \textit{i}) CNNs for segmentation of high-resolution images~\cite{scaramuzza_alphapilot}, which results in a computational complexity of more than \SI{3}{\giga OPs}~\cite{scaramuzza_alphapilot} per inference, or \textit{ii}) traditional computer vision approaches with stereo images~\cite{iros2016_winner}: both are still out of reach for nano-drones.
Kaufmann~\textit{et~al.}~\cite{kaufmann_iros2018} proposed a simpler CNN for predicting the gate's poses directly from the image, but this technique runs at only \SI{10}{\hertz} on a powerful ($\sim$\SI{13}{\watt}) Intel UpBoard and relies on a coarse map of the gate's positions.
\blue{PencilNet~\cite{pham_pencilnet_2022} is a lightweight CNN (32k parameters and 53k MAC operations per frame) for gate pose estimation trained on simulated images, addressing the sim-to-real gap with an intermediate image representation. 
From a computational/memory point of view, this CNN is suitable for real-time execution on our nano-drone.
However, the gate pose estimation is only part of a more complex pipeline to achieve gate-based navigation, which additionally requires a memory-consuming mapping of the environment and a more complex trajectory planning.
For this reason, the PencilNet CNN was demonstrated on a drone equipped with a powerful Nvidia Jetson TX2, with a power consumption 75$\times$ higher than our nano-drone's GAP8 SoC, and a high-end Intel RealSense T265 for the state estimation.}

More lightweight approaches for visual servoing have also been demonstrated aboard autonomous nano-drones, based on simple computer vision approaches, such as color segmentation, to cope with the platform's limited computational resources.
In \cite{li_visual_2020-1}, the authors used raw image data to detect and fly through monochromatic gates, while in \cite{palossi_target_2017}, a simple target-tracking algorithm for monochromatic objects was introduced. 
However, both works do not target high-speed scenarios, resulting in too limited agility for drone racing.

%% file: 03-challenge.tex
\section{The IMAV'22 Nanocopter AI Challenge} \label{sec:challenge}

\begin{figure}[t]
\centering
\includegraphics[width=1.0\linewidth]{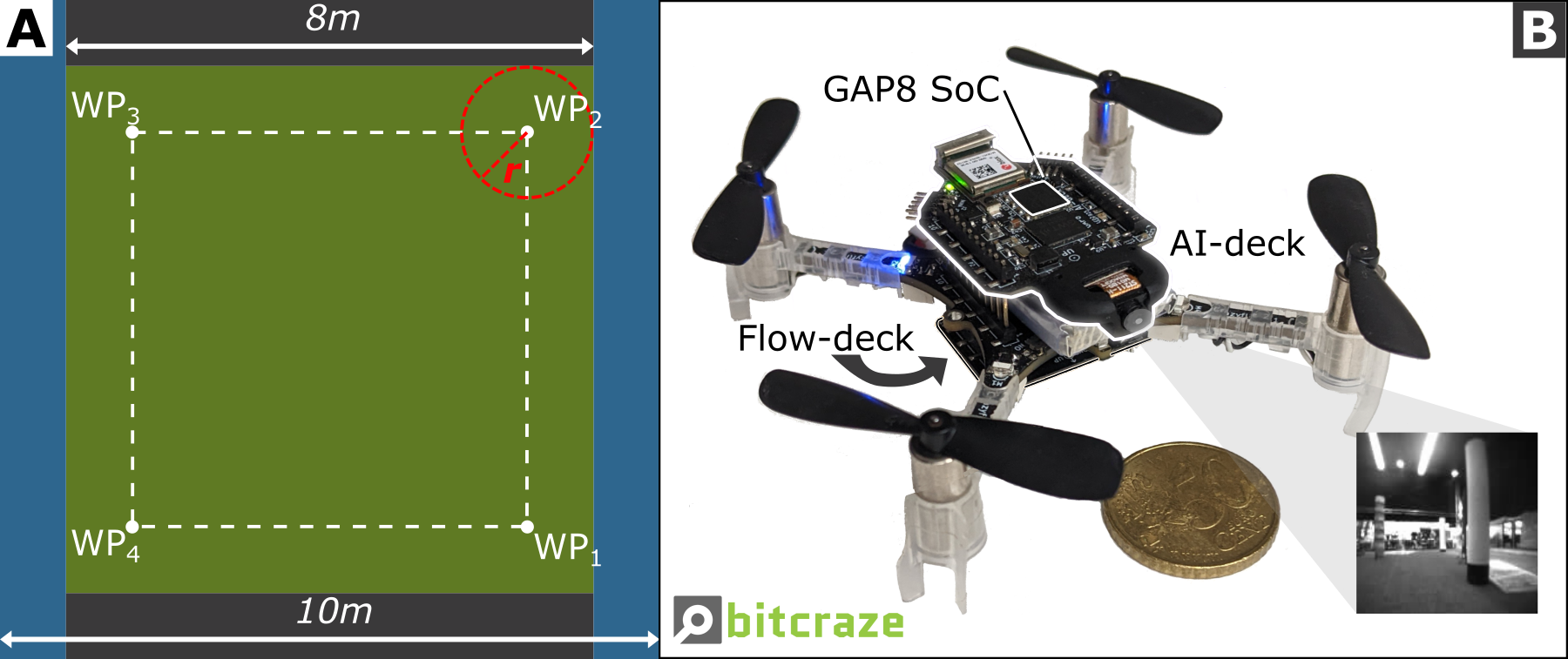}
\caption{A) the 10$\times$\SI{10}{\meter} competition arena. B) the robotic platform.}
\label{fig:mission_area}
\end{figure}

\textbf{The competition:} the metric to assess each team's score is the distance traveled within the mission area (the 8$\times$\SI{8}{\meter} green square shown in Fig.~\ref{fig:mission_area}-A) within the allotted time (\SI{5}{\minute}), employing a Crazyflie 2.1 nano-drone.
This distance is measured with a motion capture system that ignores the part of the trajectory lying outside the mission area.
Participants can choose between two coefficients of difficulty, one accounting for \textit{environmental complexity} ($\alpha_\text{env}$) and the other for \textit{computational resources} ($\alpha_\text{comp}$).
Additionally, in the arena, there are two rectangular gates through which the nano-drone can fly; every time the drone passes through a gate, the distance is increased by an additional \SI{10}{\meter}.

Participants can choose among three levels of environmental complexity: only static gates ($\alpha_\text{env}=1\times$), static gates and obstacles ($\alpha_\text{env}=5\times$), and static gates and dynamic obstacles ($\alpha_\text{env}=10\times$).
If the nano-drone uses remote computation (i.e., Wi-Fi-connected commodity laptop), the $\alpha_\text{comp}$ is $1\times$; for onboard computation, instead, it is $\alpha_\text{comp}=5\times$.
The final score is calculated as follows:

\begin{equation}
\label{eq:score}
Score = (S_\text{dist} + 10 \cdot S_\text{gates}) \cdot \alpha_\text{env} \cdot \alpha_\text{comp}
\end{equation}

where $S_\text{dist}$ is the total traveled distance and $S_\text{gates}$ is the number of times the drone passes through the gates.
Time counting is never stopped (except for battery swaps): crashing, re-flashing, rebooting, components replacements, and other flight interruptions are allowed, but they will ultimately penalize the score.
The teams can choose the starting position of the drone, but after a flight interruption, subsequent take-offs must occur near the location where the flight was interrupted.

\textbf{Robotic platform:} all teams competed with the same robotic platform comprising a COTS Bitcraze Crazyflie v2.1 drone and two additional onboard modules, as shown in Fig.~\ref{fig:mission_area}-B.
The drone is a modular \SI{27}{\gram} \SI{10}{\centi\meter} quadrotor integrating sensing, communication, and actuation subsystems with an STM32F405 MCU for sensing, state estimation (through an extended Kalman Filter) and control. 
The UAV can fly up to $\sim$\SI{7}{\minute} employing a \SI{250}{\milli\ampere\hour} battery. 
The first additional module is the Flow-deck, a small (21$\times$\SI{28}{\milli\meter}) printed circuit board (PCB) that provides additional sensors to improve the state estimation: a VL53L1x Time-of-Flight distance sensor for measuring the height from the floor and a PMW3901 low-resolution (35$\times$\SI{35}{\pixel}) optical-flow camera.

The second expansion board employed is the AI-deck, a PCB for vision-based onboard processing. 
The module embeds \textit{i}) an Himax HM01B0 low-power ($\sim$\SI{4}{\milli\watt}) QVGA monochrome image sensor camera, \textit{ii}) a NINA-W102 MCU supporting Wi-Fi and Bluetooth communications, and \textit{iii}) GAP8, an ultra-low-power (ULP) SoC capable of efficiently managing computation-intensive workloads by employing hardware-enabled (8-core cluster of processing cores) parallel programming paradigms. 
The SoC also features standard interfacing peripherals (SPI, I2C, UART) as well as different hierarchically organized SRAM memories (L1:\SI{64}{\kilo\byte} L2:\SI{512}{\kilo\byte}).

\textbf{Simulator:} the IMAV challenge organizers provided all teams with a simulator developed and released by Bitcraze, based on the open-source \textit{Webots} robot simulator.
The simulator includes a simple PID controller for the attitude control of a Crazyflie quadcopter. 
It emulates the \SI{87}{\degree} field of view of the drone's Himax camera, and it provides a photorealistic world model of the competition arena.
Several obstacles in the simulator resemble the ones of the IMAV challenge, including an orange pole; an orange gate; three panels with a width of \SI{1}{\meter}, \SI{1.5}{\meter}, and \SI{3}{\meter}, respectively; curtains; carpets; nets.

\textbf{Our strategy:} given IMAV's rules, to maximize the overall score, the nano-drone should: \textit{i}) address two concurrent complex vision-based tasks, i.e., \textit{obstacle avoidance} and \textit{gate-based navigation}; \textit{ii}) rely only on onboard computation; \textit{iii}) operate in the most challenging environment.
Relying on off-board computation can ease the computational burden. 
However, it would reduce the overall score due to the lower coefficient of difficulty and decrease the drone's reactiveness due to the higher latency of the distributed system connected via Wi-Fi.
Instead, individually addressing even one of the two tasks would represent a significant step forward in the SoA. 
Neither simulation-based obstacle avoidance nor gate-based navigation is still demonstrated in a challenging autonomous nano-drones race.
At the same time, as previously discussed, the computational and memory limitations of embedded platforms such as GAP8 still constitute a strong holdback against deploying multiple tasks simultaneously.
Therefore, for the IMAV'22 competition, we prefer to address only one of these tasks while challenging our system with fully onboard computation ($\alpha_\text{comp}=5\times$) and maximum environmental difficulty ($\alpha_\text{env}=10\times$).

To choose which task to perform, we consider two ideal nano-drones \textit{A} and \textit{B} flying at a mean speed of \SI{1.5}{\meter/\second} in a thought experiment.
\textit{A} performs obstacle avoidance in the most complex environment ($\alpha_\text{env}=10\times$), while \textit{B} addresses gate-based navigation in an obstacle-free scenario ($\alpha_\text{env}=1\times$).
In a \SI{5}{\minute} flight, the \textit{A} system would score 7500 points.
Assuming the two gates are \SI{3}{\meter} apart, the \textit{B} system would travel \SI{750}{\meter}, passing through the gates 250 times, leading to a final score of 3250 points.
Therefore, we believe that with the rule set of this competition, an autonomous nano-drone that quickly explores its surroundings and reliably avoids collision with dynamic obstacles (scenario \textit{A}) has the potential to mark a higher score than a system designed for gate-based navigation in the simpler scenario \textit{B}.

%% file: 04-navigation_policies.tex
\section{Navigation policies} \label{sec:control_policies}

\textbf{Deep Neural Network:} the neural network used in this work is derived from the open-source PULP-Dronet CNN~\cite{pulpdronetv2JETCAS}.
\blue{We keep the same network topology consisting of three consecutive residual blocks (ResBlocks).
Each ResBlock comprises a primary branch that executes two $3\times3$ convolutional layers and a parallel by-pass employing a $1\times1$ convolutional layer.}
Unlike the previous work, whose outputs consisted of a single collision probability and a steering angle, our CNN features three collision probabilities by splitting the input image horizontally in three 54$\times$\SI{162}{\pixel} left, center, and right portions of the FoV.
We exploit only simulated data to train the network, and the data collection procedure is detailed in Sec.~\ref{sec:sim_and_integration}.
After training the network, we apply fixed-point 8-bit quantization to its weights to: \textit{i}) reduce the CNN's memory footprint by $4\times$, resulting in a size of \SI{317}{\kilo\byte}, and \textit{ii}) to enable optimized 8-bit fixed-point arithmetic, resulting in a throughput of \SI{30}{frame/\second} when deployed on GAP8.
We define three navigation policies that rely on the CNN outputs: Baseline, Policy~1, and Policy~2.

\textbf{Baseline:} for each input image, the drone flies towards the direction with the lowest collision probability among the three CNN's outputs. 
In case all three probabilities are higher than a given threshold $\mathcal{T}_{h}$, the drone spins in place by \SI{180}{\degree} $\pm$ a random angle in the [\SI{0}{\degree}, \SI{30}{\degree}] range.
To generate the training labels for this model, only actual objects in the environment (i.e., panels, cylinders, walls, gates, etc.) are considered obstacles.

\textbf{Policy~1:} same as the baseline policy, with the difference that the training labels are generated considering as an obstacle also the ground outside the mission area, i.e., black and blue stripes in Fig.~\ref{fig:mission_area}-A, as well as surrounding walls.

\textbf{Policy~2:} the three probabilities of collision are trained as in Policy~1, and if at least one of them is higher than a given threshold $\mathcal{T}_{}$, the drone will behave as in the previous two policies.
Otherwise, if all probabilities of collision are lower than $\mathcal{T}_{h}$, the drone will try to reach one of the four waypoints (WPs) defined as the corners of a square inscribed in the mission area, $\text{WP}_{1,2,3,4}$ in Fig.~\ref{fig:mission_area}-A.
This \textit{WP-based navigation} mechanism leverages \textit{i}) the \textit{a priori} knowledge of the take-off point, i.e., we can choose the take-off point/orientation of the drone, for example, in $\text{WP}_1$; and \textit{ii}) the onboard visual-inertial state estimation which provides the relative position of the drone w.r.t. the take-off point.
Then, when the drone is in the WP-based navigation mode, it will try to visit the WPs in a predefined cyclical order; for example, $\text{WP}_{1,2,3,4}$ produces a counterclockwise motion, and $\text{WP}_{4,3,2,1}$ a clockwise one.
Lastly, to mark a WP as ``visited,'' we define a circular area of radius $r$ centered in each WP (in red in Fig.~\ref{fig:mission_area}-A): when the drone enters such an area, we mark the corresponding WP as visited and continue with the next one.
Whenever all three collision probabilities are higher than the threshold $\mathcal{T}_{h}$, we invert the sequence of the targeted waypoints, alternating a clockwise and counterclockwise direction.
This mechanism also serves as an escape strategy when a WP is unreachable.

\textbf{Rationale discussion}: autonomous drones flying with a priori knowledge of the environment, i.e., a map, are highly effective as we can plan for the best trajectory.
Unfortunately, the ultra-constrained resources (memory, computation, and sensors) aboard a nano-drone make this map-based navigation unfeasible~\cite{slam_offboard_crazyflie} or extremely limited~\cite{SLAMBench2}.
However, in the IMAV'22 competition, we can exploit coarse-grained information, as we know the size of the mission area, the colors of the ground, and -- up to some degree -- the shape/texture/colors of the objects thanks to the simulator provided by the organizers.
Since obstacles and gates are more likely to populate the inner part of the flying area rather than its borders, Policy~2 tries to maximize the likelihood of a straight obstacle-free path by positioning the WPs at the edges of a square. 

Therefore, ideally, we should place our four WPs precisely on the edges of the green square in Fig.~\ref{fig:mission_area}-A.
However, the unavoidable noise/drift of the simple state estimation aboard our nano-drone would often drive the vehicle outside the mission field, negatively impacting the final score.
For this reason, we introduce the WPs as the edges of an inner square within the available 8$\times$\SI{8}{\meter} mission area.

\begin{figure}[tb]
\centering
\includegraphics[width=1.0\linewidth]{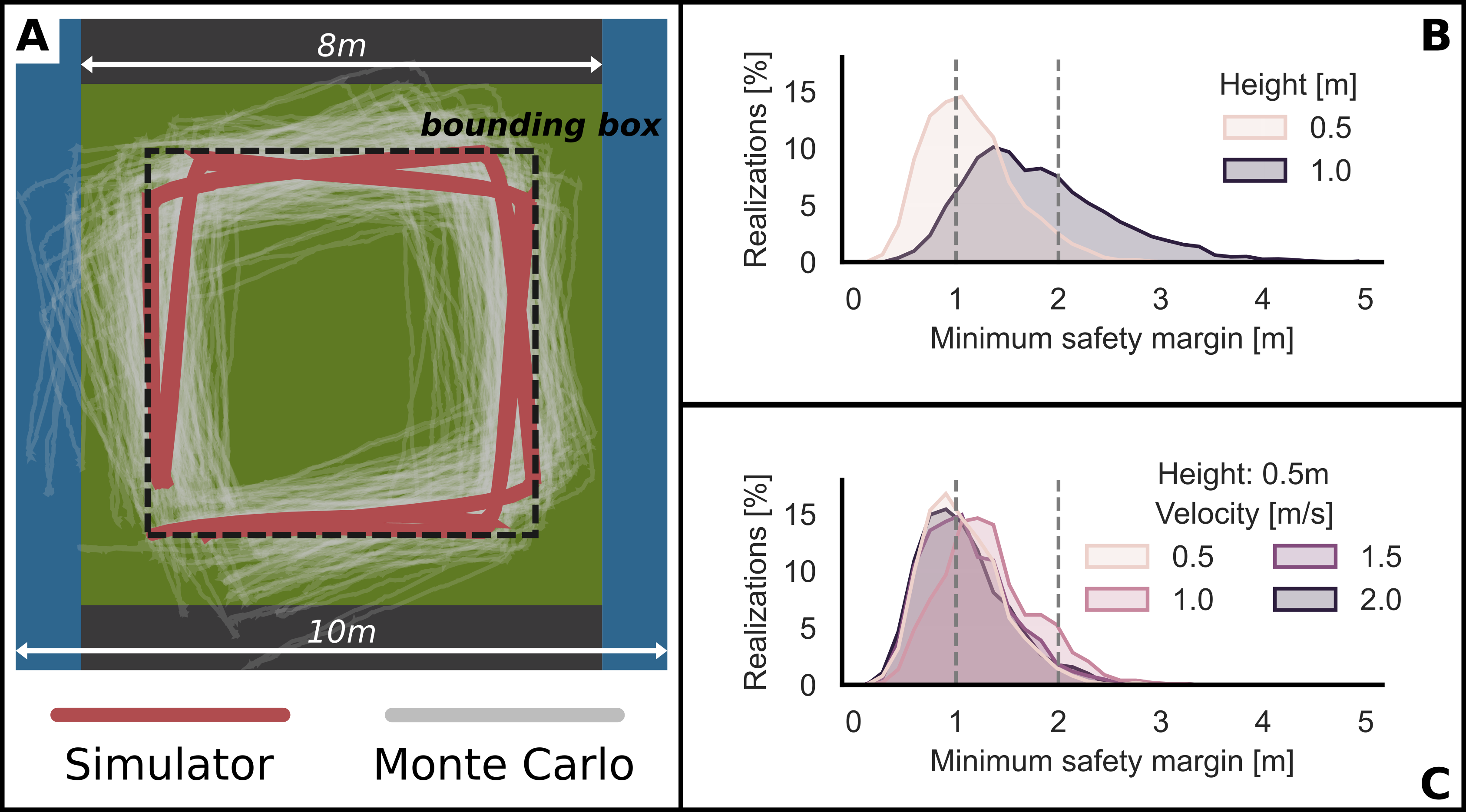}
\caption{A) 1024 simulated Monte Carlo trajectory realizations B) minimum safety margin w.r.t. height, C) minimum safety margin w.r.t speed, having fixed the height to \SI{0.5}{\meter}.}
\label{fig:state_ext_error}
\end{figure}

\textbf{State estimation analysis:} we characterize the position and orientation error of the Crazyflie's state estimation subsystem. 
This allows us to determine a margin between the WPs and the corners of the mission area that \textit{i}) keeps the drone flying close to the edges of the area and \textit{ii}) minimizes the risk that state estimation drift drives the drone out of the area.
We start by collecting real-world flight data, which we use to build a statistical model.
We rely on a setup similar to IMAV competition with artificial turf on the floor, albeit over a smaller \SI{5}{\nothing}$\times$\SI{6}{\meter} flying area.
We record the drone's movements with a motion-capture system and the drone's onboard state estimation, both at \SI{100}{\hertz}.
We perform runs of \SI{5}{\minute} each, in which the drone flies using a random walk policy while constrained through the mocap into an empty squared area -- smaller than our 5$\times$\SI{6}{\meter} arena.
We perform 16 flights: two runs for each configuration combining two flight heights, i.e., 0.5 and \SI{1.0}{\meter}, and four mean speeds, i.e., 0.5, 1.0, 1.5, \SI{2.0}{\meter/\second}.

Similarly to~\cite[Sec. 5.2.4]{siegwart2011introduction}, we quantify the state estimation error by modeling the uncertainty on incremental state estimation updates. We sample many ten-second time windows from the collected data, considering the relative pose of the drone at the beginning and the end of each window.
We measure the estimation error separately for each component of the relative pose: $x$, $y$, and yaw.
For each component, we fit a Gaussian distribution to the errors measured over many windows, then we rescale the distribution to represent the error accumulating in 1 second.

Then, in the Webots simulator, we collect eight runs, i.e., all combinations of heights and speeds, of \SI{5}{\minute} each (red lines in Fig.~\ref{fig:state_ext_error}-A).
The drone follows a squared trajectory on a \SI{6}{\nothing}$\times$\SI{6}{\meter} square, which is intuitively a safe configuration.
We invert the flown path for each lap by alternating clockwise/counterclockwise directions to ensure that any systematic errors in yaw estimation (e.g., a consistent under/over-estimation of the amount of rotation) cancel out rather than accumulate.
We use these eight flown trajectories to build a \textit{bounding box} enclosing them (dashed black line in Fig.~\ref{fig:state_ext_error}-A).
We apply the statistical error model to corrupt the ideal squared trajectory, computing 1024 Monte Carlo (M.~C.) realizations of the drone's state estimation (pale gray lines in Fig.~\ref{fig:state_ext_error}-A).
Then, only considering the portion of the M.~C. trajectories outside the bounding box, we measure the maximum distance between them and the box itself.
This measure gives us a statistical \textit{minimum safety margin} distribution that will likely constrain the flight within the desired flight area.

Fig.~\ref{fig:state_ext_error}-B shows (average on all speeds) that flying at \SI{0.5}{\meter} height requires a lower safety margin than at \SI{1.0}{\meter}, centering the distribution at \SI{1.08}{\meter} and \SI{1.56}{\meter}, respectively.
Therefore, we select \SI{0.5}{\meter} as our target height, and we focus only on this configuration in Fig.~\ref{fig:state_ext_error}-C, where we explore the four velocities. 
Among all velocities, 42\% of the 1024 realizations never exceed a safety margin of \SI{1}{\meter}, while 95.3\% satisfy a margin of \SI{2}{\meter}.
Furthermore, the median time spent outside the \SI{1}{\meter} margin is \SI{2}{\second} (95$^{th}$ percentile: \SI{31}{\second}) per realization.
Finally, of the 4.7\% of realizations that cross the margin of \SI{2}{\meter}, the median number of crossings is only one per realization (95$^{th}$ percentile: 5).
These findings confirm that positioning our WPs at the edges of a \SI{6}{\nothing}$\times$\SI{6}{\meter} square will \textit{i}) keep the drone in the desired 8$\times$\SI{8}{\meter} mission area on average for 97.6\% of the \SI{5}{\minute} run's time; and \textit{ii}) cause the drone to exceed the maximum \SI{10}{\nothing}$\times$\SI{10}{\meter} space, \textit{i.e.,} crash, only with 4.7\% probability and only once per run.
Finally, the four flight speeds analyzed do not significantly impact the safety margin requirements, which leads us to use the highest speed: \SI{2.0}{\meter/\second}.

%% file: 05-cnn_training_deployment.tex
\section{CNN training and deployment} \label{sec:sim_and_integration}

\textbf{Dataset collection:} we use the Webots simulator, \blue{shown in Fig.~\ref{fig:testing_environments}-A,} to collect a dataset of images for our CNN.
To automatically generate the labels for each image, we extended the simulator's capabilities by introducing the generation of depth and segmentation frames from the camera.
During the image collection, we use a flight height of \SI{0.5}{\meter}, see Sec.~\ref{sec:control_policies}.
Furthermore, to mimic the dynamic effects of an actual UAV flight, we add random variations for pitch, roll, and yaw in a $[-5^{\circ}; +5^{\circ}]$ range and for the height in a $[0.45; 0.55]\ m$ range. 

\begin{figure}[tb]
\centering
\includegraphics[width=1.0\linewidth]{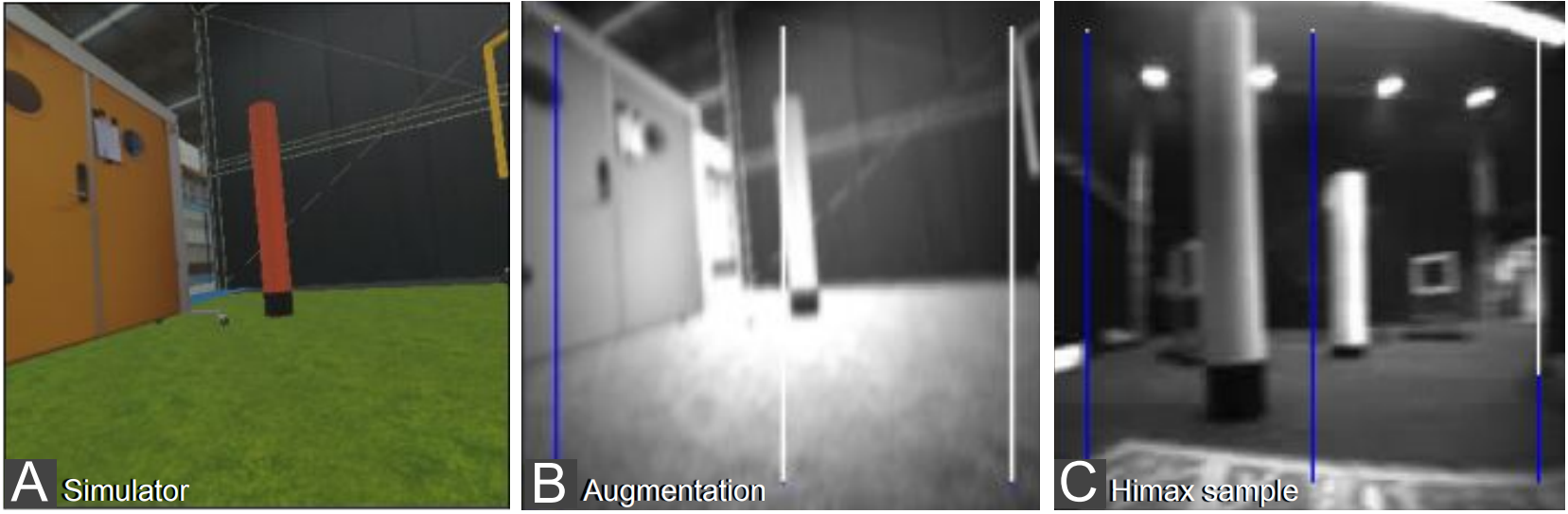}
\caption{\blue{Images: A) from the simulator, B) after augmentation, C) Himax camera sample collected in the IMAV arena. The three blue bars in images B-C) represent the three collision probabilities predicted by our network.}}
\label{fig:augmentation}
\end{figure}

\begin{figure}[tb]
\centering
\includegraphics[width=1.0\linewidth]{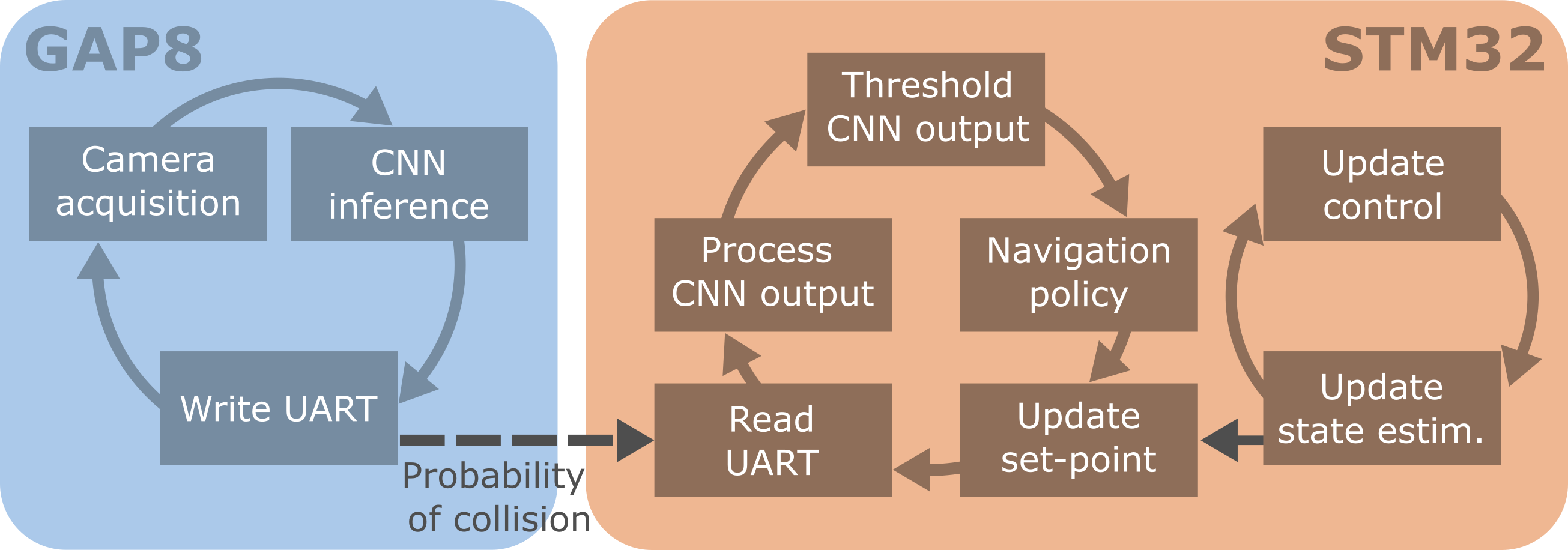}
\caption{The drone's state machines mapped on the MCUs available aboard.}
\label{fig:state_machines}
\end{figure}

We collect images with a \SI{324}{\nothing}$\times$\SI{324}{\pixel} resolution and \SI{87}{\degree} FoV, the same as our Himax camera.
For each image, we save per-pixel depth and segmentation masks, which we exploit for labeling:
we divide the camera FoV into three vertical portions of \SI{108}{\nothing}$\times$\SI{324}{\pixel}, and for each portion, we set a collision label $=1$ if the $10\%$ of the pixels belonging to obstacles has distance $\leq$\SI{2}{\meter}, 0 otherwise.
We consider all kinds of objects as obstacles, \textit{including gates} and excluding carpets.
We collected \SI{41}{\kilo\nothing} images in 3 simulated scenarios to ensure labels balancing and uniform sample distribution.
\SI{10}{\kilo\nothing} images are collected by spawning the drone randomly across the arena populated by obstacles, and \SI{21}{\kilo\nothing} images are acquired by flying around each obstacle in the simulator.
This last group comes from a "360 scan" of each obstacle, which we equally split between images with an empty background and a populated one. 
Finally, \SI{10}{\kilo\nothing} images are taken from the drone flying along a square trajectory, \SI{50}{\centi\meter} within the green flying area edges, equally split between clockwise and counterclockwise flight directions.
Eventually, we split each portion of the dataset using $70\%$ of the images for training, $10\%$ for validation, and $20\%$ for testing.

\textbf{Photometric augmentations:} as the simulated images are noise-free 3D renderings of the scene, we introduce an aggressive data augmentation pipeline to bridge the appearance gap with real-world Himax images.
We randomly perturb the simulated images to reproduce several real-world image artifacts: motion blur, Gaussian blur to simulate lens defocusing, Gaussian noise, and exposure changes (gain, gamma, dynamic range, and vignetting).
Finally, we convert images to grayscale and resize them to  162$\times$\SI{162}{\pixel}, as our Himax camera would do.
\blue{
Fig.~\ref{fig:augmentation}-A-B and the supplementary video show the result of this photometric pipeline, while Fig.~\ref{fig:augmentation}-C displays a real drone frame.
We assess the impact of the photometric augmentations on CNN accuracy by exploring all eight combinations of exposure, blur, and noise augmentations. 
To evaluate accuracy, we collected 2200 real-world images from our 5$\times$\SI{6}{\meter} indoor flying arena, shown in Fig.~\ref{fig:testing_environments}-B. 
The results in Tab.~\ref{tab:ablation_augmentation} shows that enabling all augmentations scores the highest accuracy, surpassing the non-augmented model by 17\%.
}

\begin{table}
\centering
\footnotesize
\caption{Neural network accuracy by photometric augmentation method: Noise (N), Blur (B), and Exposure (E).}
\label{tab:accuracy}
\color{blue}
\begin{tblr}{
  width = \linewidth,
  colspec = {Q[100]Q[90]Q[110]Q[90]Q[90]Q[90]Q[90]Q[90]Q[140]},
  column{even} = {c},
  column{3} = {c},
  column{5} = {c},
  column{7} = {c},
  column{9} = {c},
  hline{1,3} = {-}{0.08em},
}
\textbf{Aug.} & \textbf{N} & \textit{\textbf{None}} & \textbf{B} & \textbf{B+E} & \textbf{N+B} & \textbf{E} & \textbf{N+E} & \textbf{N+B+E} \\
\textbf{Acc.} & 55\%       & 56\%                   & 58\%       & 62\%         & 64\%         & 68\%       & 69\%         & 72\%           
\end{tblr}
\label{tab:ablation_augmentation}
\end{table}

\textbf{Navigation policies implementation:} Fig.~\ref{fig:state_machines} shows the state machines mapped on the two MCUs available aboard our UAV.
GAP8 implements a loop for \textit{i}) image acquisition, \textit{ii}) CNN inference, and \textit{iii}) UART transmission of the 3 CNN collision probability outputs, i.e., $\{\text{left},\text{center},\text{right}\}$, to the STM32.
The STM32 instead implements two loops for high and low-level control of the drone, respectively.
The high-level control loop \textit{i}) reads the CNN's output from UART, \textit{ii}) converts the 8-bit fixed-point CNN's outputs ([0,255] range) to \texttt{float32} numbers ([0,1] range) and applies a low pass filter, \textit{iii}) thresholds the three probabilities of collision to$\mathcal{T}_{h}=0.7$, getting values $\in\{0,1\}$, \textit{iv)} applies the navigation policy to calculate the next set point, and \textit{v)} updates the low-level controller.
Instead, the low-level control loop, running at \SI{100}{\hertz}, updates the state estimation through the eKF and applies a cascade of PID controllers to reach the set point pushed by the high-level control loop.

All three navigation policies described in Sec.~\ref{sec:control_policies} output a \{speed, yaw rate\} tuple as a set point.
For all policies, the forward speed is inversely proportional to the \textit{center} collision probability. 
The yaw rate is set to -90 or +90 \SI{}{\deg/\second} when the right or the left collision probabilities get over the threshold $th$, respectively.
If all three collision probabilities exceed the threshold $\mathcal{T}_{h}$, the drone spins in place as described in Sec.~\ref{sec:control_policies}.
Conversely, the three policies act differently when all collision probabilities are zero: Baseline and Policy~1 command the drone to fly in a straight line, while Policy~2 activates the WP-based navigation mode, heading the drone to the next WP. 

%% file: 06-results.tex
\section{Experimental results} \label{sec:results}

\begin{figure}[tb]
\centering
\includegraphics[width=1.0\linewidth]{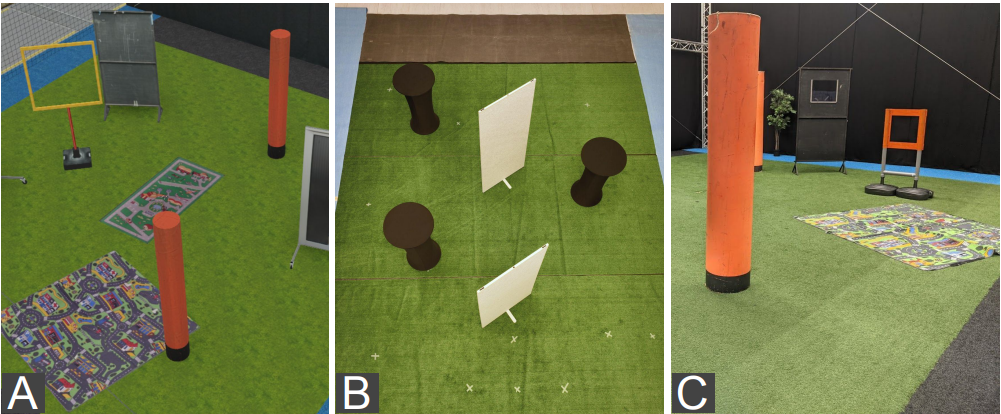}
\caption{Three environments: A) the \textit{Webots} simulator, B) our indoor \SI{5}{\nothing}$\times$\SI{6}{\meter} arena, and C) the 8$\times$\SI{8}{\meter} competition arena.}
\label{fig:testing_environments}
\end{figure}

\begin{figure}[tb]
\centering
\includegraphics[width=1\linewidth]{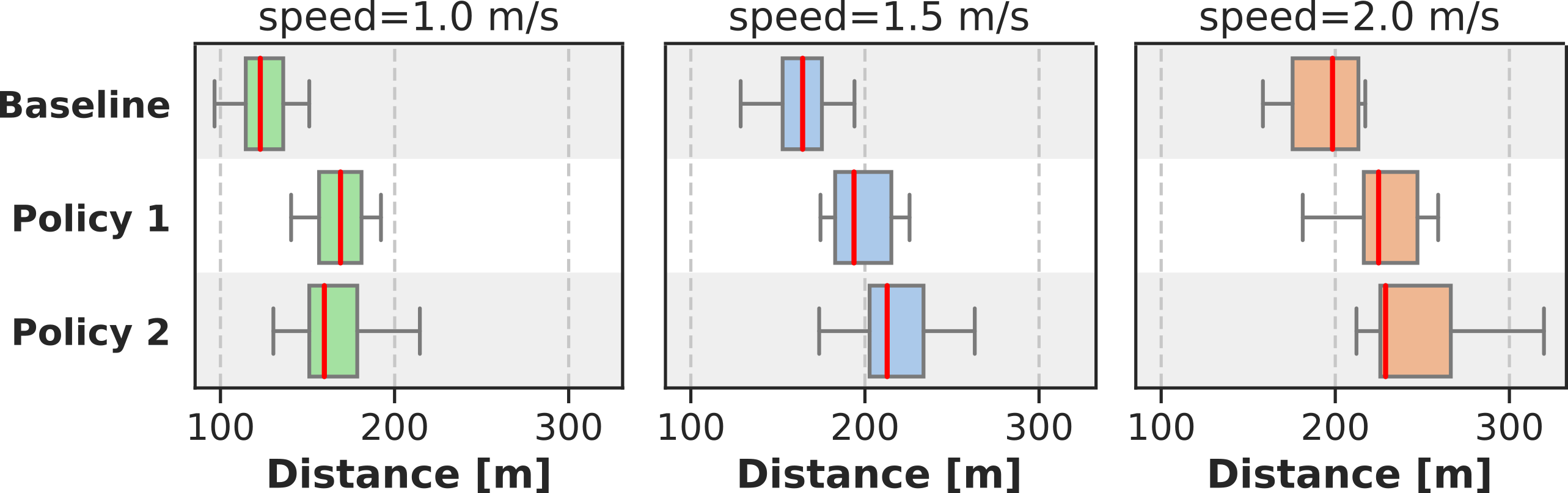}
\caption{Median traveled distance (10 runs) in \SI{5}{\minute} flights on the simulator of our three navigation policies at 1.0, 1.5, and \SI{2.0}{\meter/\second}.}
\label{fig:results_sim}
\end{figure}

\subsection{In-simulator evaluation} \label{sec:sum_results}

We evaluate our three navigation policies with the Webots-based simulator, sample picture shown in Fig.~\ref{fig:testing_environments}-A.
We rely on an ideal flight controller inside the simulator, meaning \textit{i}) no state-estimation drift and \textit{ii}) exploiting depth maps for ideal obstacle avoidance.
We tested nine configurations combining the three control policies introduced in Sec.~\ref{sec:control_policies} and at three speeds ($1, 1.5$, and \SI{2}{\meter/\second}).
For all the tests, we used seven obstacles: two orange poles, two gates, and three panels with a width of $1, 1.5$, and \SI{3}{\meter}, respectively.
We run ten experiments of \SI{5}{\minute} each for each configuration, varying the obstacles' initial position but keeping it consistent among the nine configurations.
To emulate the dynamic obstacles of the challenge, we move one obstacle every \SI{30}{\second}.
We compare the policies based on the distance traveled on the green turf.

Fig.~\ref{fig:results_sim} shows that the baseline policy (which does not consider the floor outside the green turf as an obstacle) is consistently worse than the others, spending $16\%$ of the time outside the mission area. 
Conversely, Policy~1 and 2 detect the external area as an obstacle, spending only $1\%$ and $0.8\%$ of the time on the external area, respectively.
Both Policy~1 and 2 benefit from higher flight speeds due to the simulator's perfect sensing preventing collisions.
At the lowest speed configuration of \SI{1}{\meter/\second}, Policy~2 achieves a slightly lower median distance than Policy~1, \SI{165}{\meter} vs. \SI{169}{\meter}.
Instead, for higher speed configurations, the median distance of Policy~1 is slightly lower than Policy~2: \SI{217}{\meter} and \SI{246}{\meter} at \SI{1.5}{\meter/\second}, and \SI{193}{\meter} and \SI{225}{\meter} at \SI{2}{\meter/\second}, respectively.
Since both Policy~1 and Policy~2 score a similar traveled distance, we push forward our analysis by deploying and testing both of them in the field.

\subsection{\blue{State-of-the-Art comparison}} \label{sec:soa_comparison}

\blue{
Since our CNNs are inspired by the PULP-Dronet~\cite{pulpdronetv2JETCAS}, as well as many other teams at the competition, we present a thorough comparison of our system against a \textit{vanilla} PULP-Dronet and a fine-tuned version of it.
Given the different outputs between our CNN and the original one, we adapt our navigation Policy~1 (no WPs) to the PULP-Dronet baselines.
We use the PULP-Dronet steering output, as in its original implementation, by converting it in a yaw-rate for the controller in case of a probability of collision lower than 0.7, while the drone rotates 180\textdegree otherwise.
For the fine-tuned version, we use the same dataset collected in simulation to train our models.
The single collision label of PULP-Dronet is matched to the central probability of our models, while the steering angle reflects the direction with the lower probability of collision among the three probabilities: left, center, and right.
}

\begin{figure}[tb]
\centering
\includegraphics[width=1\linewidth]{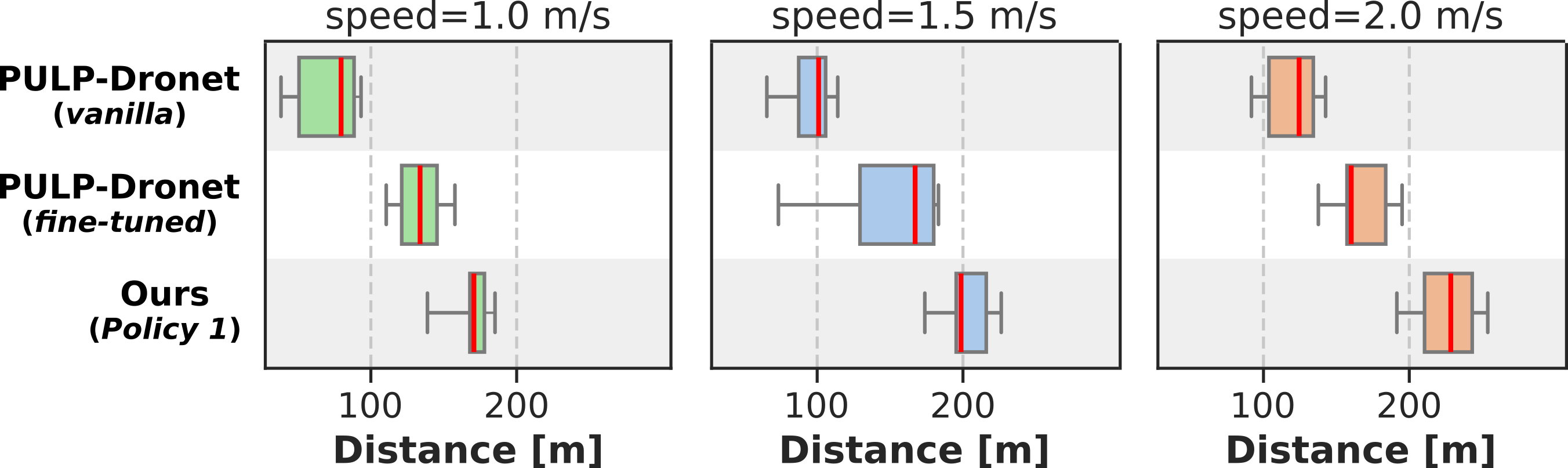}
\caption{
    \blue{
    Median traveled distance (10 runs) in \SI{5}{\minute} flights on the simulator comparing our work to the SoA PULP-Dronet~\cite{pulpdronetv2JETCAS} at 1.0, 1.5, and \SI{2.0}{\meter/\second}.
    }
}
\label{fig:x_result_soa_comparison}
\end{figure}

\blue{
The results in Fig.~\ref{fig:x_result_soa_comparison} are collected in simulation, where we run ten 5-\SI{}{\minute} tests for each model at three target flight speeds (i.e., 1, 1.5, and \SI{2}{\meter/\second}).
The vanilla PULP-Dronet marks the lowest score in all configurations, 84–114\% less than our system, while the fine-tuned PULP-Dronet reduces this gap with a performance reduction of 28\%–43\%, depending on the flight speed. 
Our network scores a median distance traveled of 171, 199, and \SI{228}{\meter} at 1, 1.5, and \SI{2}{\meter/\second}, respectively, showing the benefit of our architectural changes and training methodology.
}

\subsection{In-field evaluation} \label{sec:in_field_test}

\begin{figure}[tb]
\centering
\includegraphics[width=\linewidth]{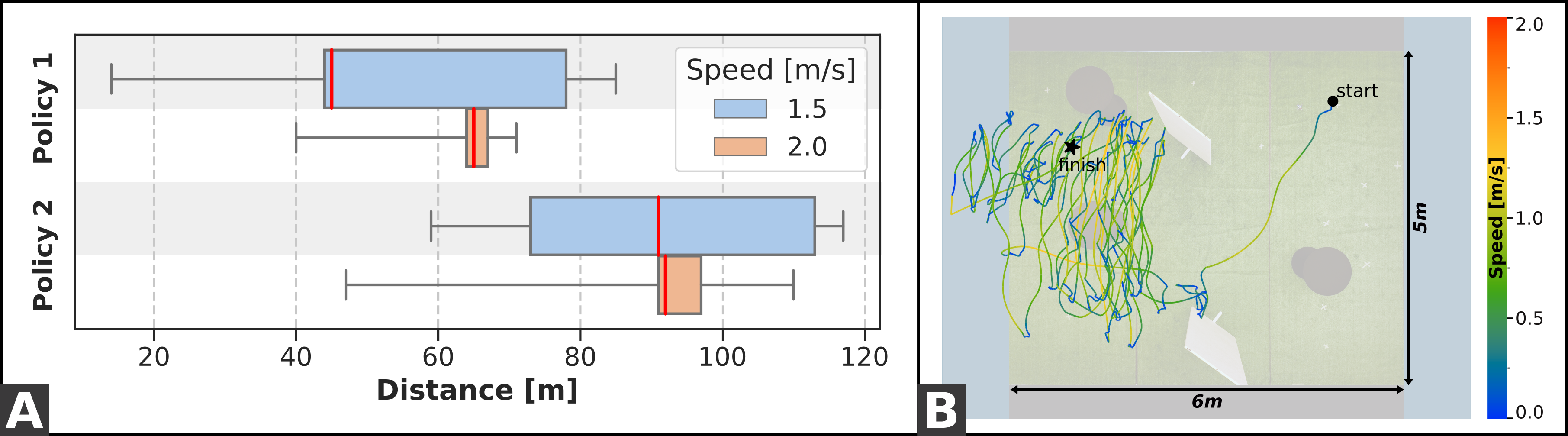}
\caption{In-field testing, A) median traveled distance (5 runs) in \SI{5}{\minute} flights of our two navigation policies at 1.5 and \SI{2.0}{\meter/\second}. B) Sample run of Policy~2 with a $v_{target}$ = \SI{1.5}{\meter/\second}, scoring \SI{117}{\meter} of traveled distance.}
\label{fig:results_tii}
\end{figure}

We evaluate Policy~1 and Policy~2 in our 5$\times$\SI{6}{\meter} indoor flying arena, testing them with two target speeds, 1.5 and \SI{2}{\meter/\second}, resulting in four configurations.
For each configuration, we perform 5 runs of \SI{5}{\minute} each.
We use three black cylinders and two white panels as obstacles, as shown in Fig.~\ref{fig:testing_environments}-B. 
We replicate the competition's conditions, e.g., periodically moving the dynamic obstacles (every $\sim$\SI{30}{\second}), and without stopping the time count if the drone crashes.
As shown in Fig.~\ref{fig:results_tii}-A, Policy~1 scores a median 45 and \SI{65}{\meter}, while Policy~2 marks a median 91 and \SI{92}{\meter}, while flying at 1.5 and \SI{2}{\meter/\second}, respectively.
\blue{The improved performance of Policy~2 derives from the \textit{WP-based navigation}, which provides the drone with boundaries, limiting the time spent in potentially dangerous areas outside the mission field.}
A sample run of Policy~2 is shown in Fig.~\ref{fig:results_tii}-B.
For the IMAV competition, we choose to use the Policy~2 at \SI{1.5}{\meter/\second} for the first run (conservative) and push our system to the \SI{2}{\meter/\second} limit in our second attempt.

\subsection{The nanocopter AI challenge} \label{sec:competition}

The arena for the competition was a 10$\times$\SI{10}{\meter} area, represented in Fig.~\ref{fig:mission_area}-A, where only the 8$\times$\SI{8}{\meter} green surface is considered for the final score.
The arena had ten objects (Fig.~\ref{fig:testing_environments}): four orange poles, two black panels, two flags, and two gates.
The competition allows each team to perform two runs of \SI{5}{\minute} each, having only the best one considered in the final leaderboard.
Six teams participated, and the video recording is available online\footnote{\url{http://youtu.be/WaDU4I2TImA}}. 
For each team, Tab.~\ref{tab:leaderboard} summarizes the distance traveled, the number of gates passed, and the final score computed according to Eq.~\ref{eq:score}. 

The last team achieved a traveled distance of \SI{9}{\meter} in an obstacle-free environment while employing an off-board (i.e., WiFi-connected laptop) color segmentation algorithm.
The 5\textsuperscript{th} ranked team tackled an obstacle avoidance task using an off-board CNN for monocular depth estimation.
They exploited the depth to control the drone's forward speed and steering angle commands, ultimately reaching a traveled distance of \SI{67}{\meter}.
The 4\textsuperscript{th}-classified team proposed a system that passed through four gates, i.e., the highest number of traversed gates among all teams. 
They tackled gate-based navigation with an off-board color segmentation visual pipeline. 
However, their final traveled distance of \SI{31}{\meter} was penalized by numerous crashes, wasting precious time for resuming the flight.

\blue{
The team ranked 3\textsuperscript{rd} used an onboard vanilla PULP-Dronet for obstacle avoidance, which led to a  conservative $\sim$\SI{0.1}{\meter/\second} average speed on their best 5-\SI{}{\minute} run (\textit{run~1}).
Nevertheless, they managed to pass through one gate while crashing multiple times (up to five in \textit{run~2}), resulting in a traveled distance of \SI{29}{\meter}.
The second classified team exploited only onboard computation in a static obstacles environment. 
Their navigation strategy limited the exploration to a small obstacle-free part of the environment, leading to a total distance of \SI{81}{\meter}.
}

\begin{figure}[tb]
\centering
\includegraphics[width=\linewidth]{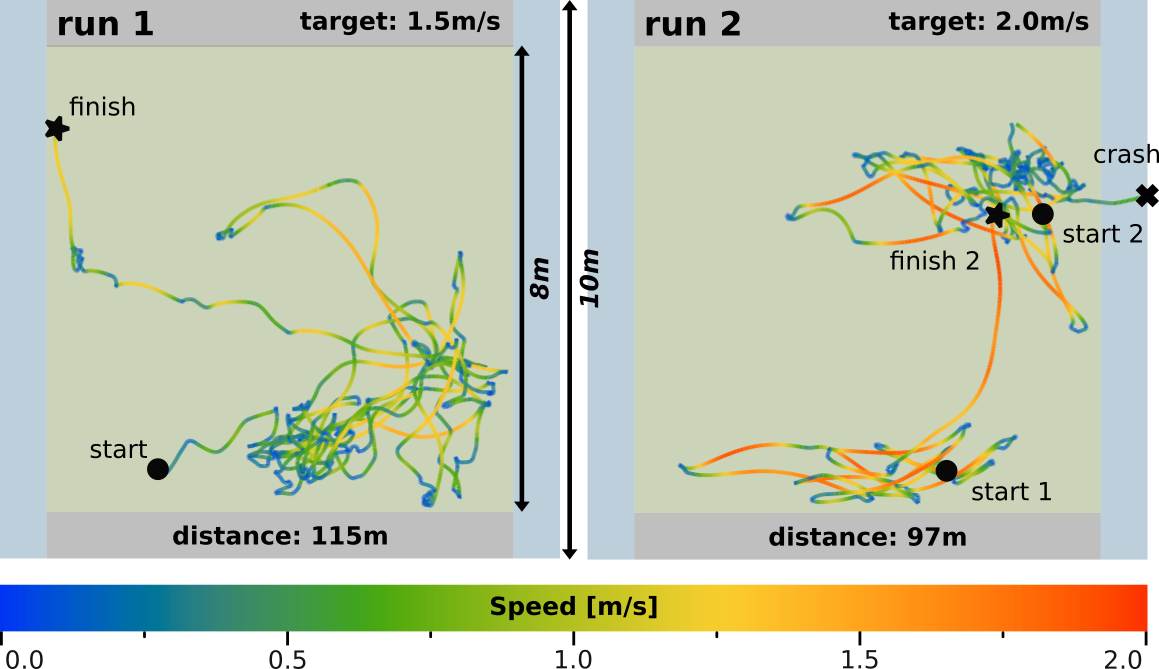}
\caption{Trajectories of our two runs during the IMAV'22 race: run 1 ranked our team $1^{st}$ at the ``Nanocopter AI challenge.''}
\label{fig:race_traj}
\end{figure}

We tackled the most challenging scenario, i.e., dynamic obstacles, with only onboard computation (i.e., no WiFi-connected laptop), while leveraging our Policy~2, described in Sec.~\ref{sec:control_policies}.
In our two runs, we set a maximum target speed of 1.5 \SI{2.0}{\meter/\second}, which resulted in a traveled distance of 115 and \SI{97}{\meter}, respectively.
During the first and best run, we never crashed for the entire 5-minute flight, resulting in the winning score of 5750 points, i.e., almost 3$\times$ more than the second-classified team.
Fig.~\ref{fig:race_traj} summarizes our two runs, reporting a top-view of the arena where flight trajectories are shown with their mean average speed.

\begin{table}
\centering
\caption{The final leaderboard of the ``Nanocopter AI challenge.''}
\label{tab:leaderboard}
\color{blue}\begin{tabular}{ l c c c c}
\toprule
\textbf{Team} & \textbf{Processing} & \textbf{Obstacles}  & \textbf{[m]~/~Gates} & \textbf{Score}\\
\midrule
\textbf{PULP (ours)} & Onboard & Dynamic & 115~/~0 & 5750 \\
Black Bee Drones & Onboard & Static & 81~/~0 & 2015 \\
SkyRats & Onboard & Dynamic & 29~/~1 & 1945 \\
CVAR-UPM & Off-board & Dynamic & 31~/~4 & 702 \\
CrazyFlieFolder & Off-board & Dynamic & 67~/~0 & 666 \\
RSA & Off-board & No & 9~/~0 & 9 \\
\bottomrule
\end{tabular}
\end{table}

%% file: 07-conclusion.tex
\section{Conclusion and future work} \label{sec:conclusion}

We present the deep learning framework for visual-based autonomous navigation aboard nano-drones, winning the ``IMAV'22 Nanocopter AI Challenge'' drone race. 
Our system combines a CNN for obstacle avoidance, trained only in simulation, a sim-to-real mitigation strategy, and a navigation policy, defining the drone's control state machine.
Our system scored \SI{115}{\meter} of traveled distance at the competition while coping with static and dynamic obstacles.
\blue{
Since we focused on the most challenging obstacle avoidance task while leaving out the gate-based navigation task, future work will address developing lightweight perception modules for both tasks.
Nevertheless, our system marks the SoA being the first example of an autonomous nano-drone completing its mission (\SI{5}{\minute} flight, at \SI{1.5}{\meter/\second} with no crashes) in a challenging never-seen-before race environment.
Additionally, any bigger robot can exploit our lightweight yet accurate and reactive perception, freeing a vast amount of computational resources and memory that can be allocated to tackle additional complex tasks.
By sharing our knowledge, we foster the research by providing a solid foundation for future development in the newborn field of nano-drone racing.
}
